\documentclass[a4paper,fleqn]{cas-dc}

\usepackage[numbers]{natbib}
\usepackage{algorithm}
\usepackage{algorithmic}
\usepackage{multirow}
\usepackage{threeparttable}
\usepackage{color}
\usepackage{subfigure}
\usepackage{caption}
\usepackage{capt-of}

\def\tsc#1{\csdef{#1}{\textsc{\lowercase{#1}}\xspace}}

\tsc{WGM}
\tsc{QE}
\tsc{EP}
\tsc{PMS}
\tsc{BEC}
\tsc{DE}

\begin{document}
\let\WriteBookmarks\relax
\def\floatpagepagefraction{1}
\def\textpagefraction{.001}

\shorttitle{Infer-AVAE}
\shortauthors{Yadong Zhou et~al.}

\title [mode = title]{Infer-AVAE: An Attribute Inference Model Based on Adversarial Variational Autoencoder}                      
\author[1]{Yadong Zhou}[type=editor,
                        auid=000,bioid=1]
\ead{ydzhou@xjtu.edu.cn}

\author[1]{Zhihao Ding}[type=editor,
                        auid=000,bioid=1]
\ead{tommydzh@stu.xjtu.edu.cn}

\author[1]{Xiaoming Liu}[type=editor,
                        auid=000,bioid=1]
\cormark[1]
\ead{xm.liu@xjtu.edu.cn}

\author[1]{Chao Shen}[type=editor,
                        auid=000,bioid=1]
\ead{chaoshen@xjtu.edu.cn}

\author[2]{Lingling Tong}[type=editor,
                        auid=000,bioid=1]
\ead{tongling300@sina.com}

\author[1,3]{Xiaohong Guan}[type=editor,
                        auid=000,bioid=1] 

\ead{xhguan@xjtu.edu.cn}

\address[1]{MOE Key Lab for Intelligent Networks and Network Security, Xi'an Jiaotong University.}
\address[2]{National Computer network Emergency Response technical Team.}
\address[3]{Center for Intelligent and Networked Systems and TNLIST Lab, Tsinghua University.}
\cortext[cor1]{Corresponding author.}
\begin{abstract}
User attributes, such as gender and education, face severe incompleteness in social networks. 
In order to make this kind of valuable data usable for downstream tasks like user profiling and personalized recommendation, attribute inference aims to infer users' missing attribute labels based on observed data. 
Recently, variational autoencoder (VAE), an end-to-end deep generative model, has shown promising performance by handling the problem in a semi-supervised way. 
However, VAEs can easily suffer from over-fitting and over-smoothing when applied to attribute inference.
To be specific, VAE implemented with multi-layer perceptron (MLP) can only reconstruct input data but fail in inferring missing parts. 
While using the trending graph neural networks (GNNs) as encoder has the problem that GNNs aggregate redundant information from neighborhood and generate indistinguishable user representations, which is known as over-smoothing.
In this paper, we propose an attribute \textbf{Infer}ence model based on \textbf{A}dversarial \textbf{VAE} (Infer-AVAE) to cope with these issues. 
Specifically, to overcome over-smoothing, Infer-AVAE unifies MLP and GNNs in encoder to learn positive and negative latent representations respectively.
Meanwhile, an adversarial network is trained to distinguish the two representations and GNNs are trained to aggregate less noise for more robust representations through adversarial training.
Finally, to relieve over-fitting, mutual information constraint is introduced as a regularizer for decoder, so that it can make better use of auxiliary information in representations and generate outputs not limited by observations. We evaluate our model on 4 real-world social network datasets, experimental results demonstrate that our model averagely outperforms baselines by 7.0$\%$ in accuracy.
\end{abstract}


\maketitle

\section{Introduction}

With the popularity of online social networks, more and more users provide demographic information such as gender, education, and location on their accounts as their personal identification. These user attributes, on the other hand, is a kind of  important social network data supporting applications like personalized recommendation, user profiling, community detection, etc. In real-world social networks, however, user attributes are severely incomplete. For example, only 40\% of Facebook users provide their employers ~\cite{employers} and 16\% of Twitter users provide their home cities~\cite{home}. This kind of situation makes attribute data suffers from severe sparsity and hard to be made use of. Take online recommendation as an example, without the gender information, advertisers may recommend products for female to male users, which is a waste of resources. Aiming at inferring the missing user attributes, attribute inference is important for online services to take full advantage of this kind of valuable data. 

Inspired by the phenomenon of homophily~\cite{homophily,homo} and easy access to users' social connections in social networks, researchers have been attempting to infer attributes through social connections. Existing works used to regard attribute inference as a classification problem and classify users to different attribute labels using features extracted from network structure~\cite{Ego,EdgeExplain,
Mobi}. But these methods suffered from designing features manually which is often ineffective in a different dataset, and they are usually computationally complicated.

\begin{figure}
	\centering
		\includegraphics[scale=.45]{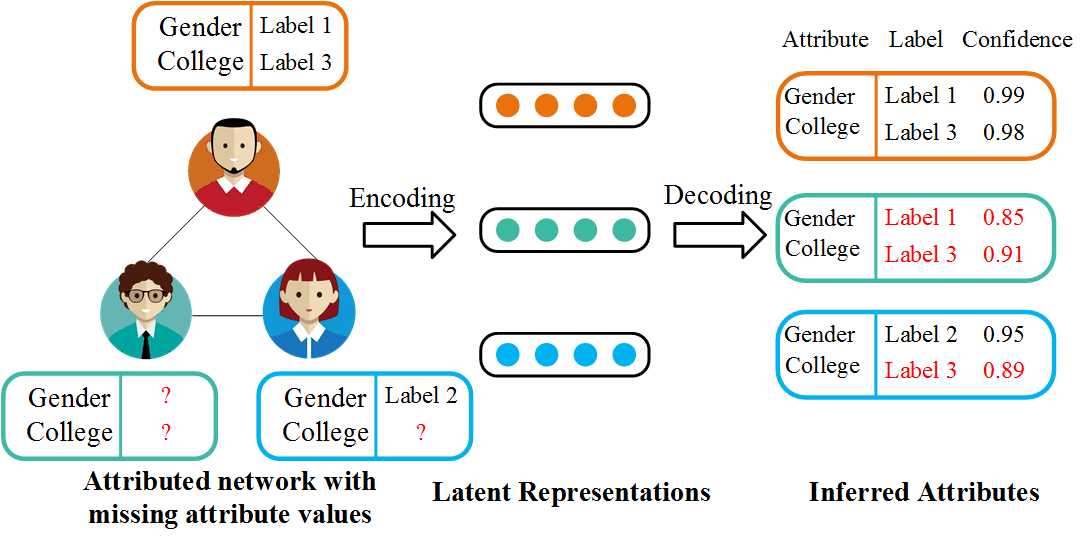}
\caption{\textbf{An example of attribute inference on social network based on variational autoencoder.} VAE takes incomplete attribute data as input and learn latent representation for each user through encoder. Decoder then reconstructs input data with missing attribute values inferred according to the learned representations.}
\label{fig:illus}       
\end{figure}

\begin{figure}[tp]
	\centering
		\includegraphics[scale=.4]{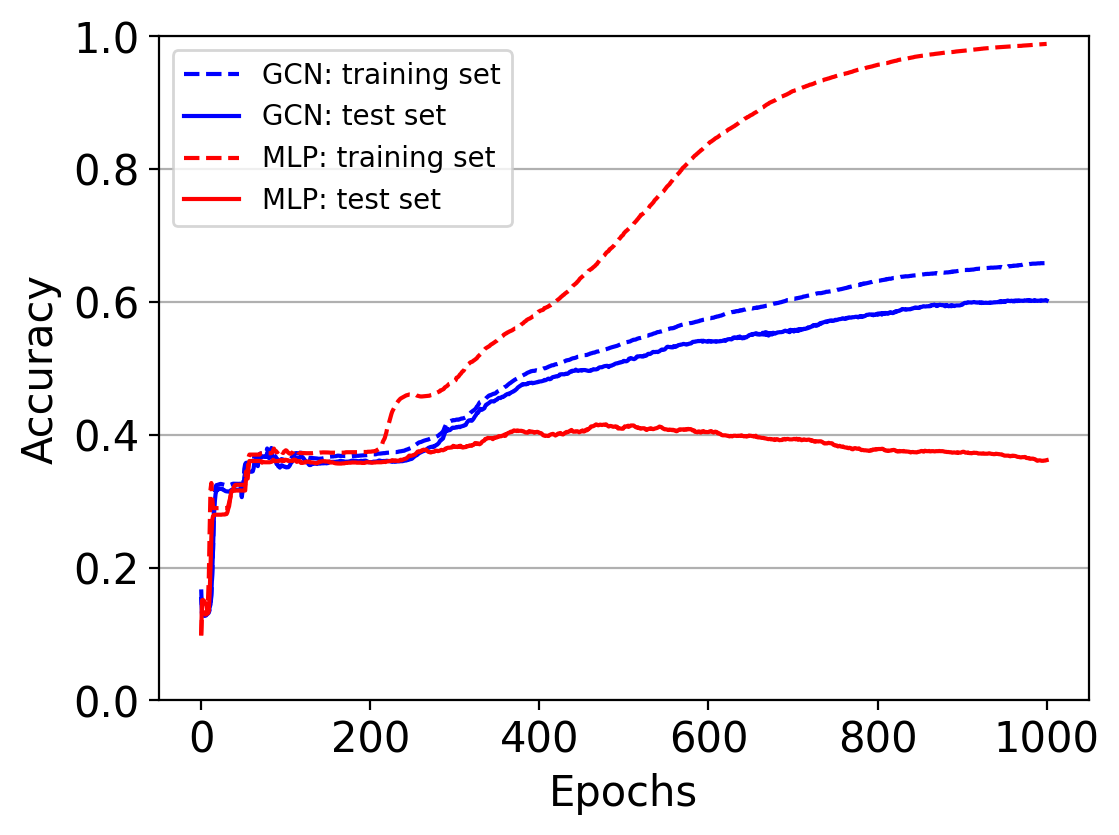}
\caption{\textbf{Performance of VAEs implemented with MLP and GNNs on both the training set and test set of fb-CMU.} VAE using MLP as the encoder suffers from severe over-fitting, it can fit the training data well while fails in test set. On the contrary, GNNs-based-VAE achieves performance upsurge in test set while has trouble in converging in training data due to over-smoothing. The details of experiment's implementation will be given in Section 5.1.}
\label{fig:mlp_gcn}       
\end{figure}

Recently, variational auto-encoder, which is an end-to-end deep generative model, has been applied to regard attribute inference as a semi-supervised problem~\cite{CAN,CCEM}. As shown in Figure \ref{fig:illus}, VAE embeds sparse attribute data into latent representation through encoder for each user and reconstructs input data from decoder with missing attribute values inferred.
Using VAE for attribute inference has the benefit that the learned latent representation can capture the distribution of attribute data ~\cite{AVAE} without extra human interference and relieve attribute inference from a classification problem. Such encoder-decoder structure improves efficiency dramatically and is more suitable for today's social networks with a bulk of users and attributes.
But VAE's performance depends on the expressiveness of the learned latent representations heavily. Through experiments, we find that VAE shows unsatisfactory performance in attribute inference.
As shown in Figure~\ref{fig:mlp_gcn},
when we use simple MLP as VAE's encoder, the model can fit the input data well (achieve high accuracy in the training set), but it contain little useful information for inferring missing values, which leads to poor performance in the test set. For more expressive representations, some researchers~\cite{CAN,HGAT} tried to adopt graph neural networks as encoder to aggregate extra attribute information from the neighborhood.
Even though VAEs implemented with GNNs performs much better than the ones using MLP, GNNs still have limitations that hinder further performance improvement. The most common one is over-smoothing~\cite{deeper}, which indicates that GNNs will generate indistinguishable representations after several layers' message passing based on graph structure~\cite{chen2020measuring}. 
This issue become more severe in attribute inference where users have multiple attributes types and labels. 
For example, users graduated from the same college will connect densely with each other while their other attributes like gender and employer may not be the same. 
When GNNs generate "similar" user representations, as the results shown in Figure~\ref{fig:mlp_gcn}, the decoder of VAE has difficulty in reconstructing the input data and converging~\cite{dropedge}, let alone inferring missing data. Inferring attributes using these indistinguishable representations is sure to be a disaster. Due to the problem aforementioned, simple combination of VAE and GNNs cannot generate robust latent representations for accurate attribute inference either. 

In general, current VAE-based models have the problem of over-fitting and over-smoothing when they are applied to attribute inference. In this paper, we propose an attribute inference model based on adversarial VAE (Infer-AVAE) to alleviate the above issues. Inspired by Generative Adversarial Network (GAN)~\cite{GAN}, our model equips VAE with adversarial network to relieve over-smoothing. As shown in Figure \ref{fig:frame}, Infer-AVAE first unifies MLP and GNNs in the encoder to learn dual latent representations. To be specific, MLP layers of the encoder only embed the observed attributes data into mid latent representations which contain limited but valid data of each user. GNN layers are then adopted to converge mid latent representations according to social connections and learn user latent representations. In contrast, user latent representations generated from GNN layers contain extra information for inference but can be noisy. After having dual latent representations, we train an adversarial network to leverage the difference between the two representations and regularize GNN layers to aggregate information of less noise through adversarial training. What's more, apart from focusing on learning latent representations like existing works, we specially design a regularization term, which is called mutual information constraint, to regularize the decoder and relieve over-fitting. The constraint evaluates mutual information between the dual latent representations, and encourages decoder to make more use of extra information. In this way, the decoder is encouraged to generate diverse outputs in the process of inference rather than reconstructing the inputs only.

Our contributions can be summarized as follows:
\begin{itemize}
    \item We propose a new model, Infer-AVAE, which unifies VAE with adversarial network under one framework to relieve GNNs from over-smoothing and improve the performance of VAE in attribute inference. 
    \item Unlike existing works focusing on learning representations only, we specifically design a regularization term for the decoder to alleviate over-fitting by estimating mutual information.
    \item We evaluate the proposed model over 4 real-world social networks datasets. The results demonstrate that our model not only achieves performance improvements compared to baseline methods but also gets robust results under different settings of label sparsity.
\end{itemize}

The rest of this paper is organized as follows. In Section 2, the related work was introduced. In Section 3, we give the formal problem formulation of attribute inference. In Section 4, we introduce our model, including the framework and training strategy, in detail. In Section 5, the results of the experiments are presented and discussed. Finally, we ends the paper with conclusions and acknowledgements.

\section{Related Work}

\subsection{Attribute Inference}

There have a lot of methods been studied inferring attributes via social connections~\cite{LP,BLA,EdgeExplain,Ego,CAN,HGAT}. Label propagation (LP) is a semi-supervised method for spreading node attributes (labels) based on link structure ~\cite{LP}. BLA~\cite{BLA} makes use of interaction between users' attributes and links and iteratively addresses the problems of link prediction and attribute inference to promote both performances. To further explore the relationship between attributes and social links, some methods~\cite{EdgeExplain,Ego} discriminate attribute types and the corresponding links, and thus significantly outperform label propagation when inferring user attributes with multiple types. But these approaches need designing hand-crafted features which are usually partial for complicated social networks and inefficient. 

Recently, inspired by the success of graph representation learning on graph-related tasks, researchers~\cite{CAN,HGAT} used graph neural networks to learn latent representation for users and infer attributes according to the representations.
For example, CAN~\cite{CAN} gets user embedding and attribute embedding collaboratively from relation data and attribute data using GCN, and the model can solve various problems on graphs including attribute inference. HGAT~\cite{HGAT} incorporates attention mechanism in heterogeneous graph neural network and solves the problem of attribute inference (user profiling) with the help of relationships between multiple kinds of entities.

However, the learned representations usually contains too much irrelevant information and suffers from severe over-smoothing~\cite{deeper}, which impedes these models from making accurate attribute inference in social networks.

\subsection{Variational Auto-encoders}

Variational Auto-encoders~\cite{repara, CFVAE,NVAE} are a kind of deep generative model that is comprised of an encoder and a decoder.
CAN~\cite{CAN} uses Graph Convolution Networks (GCN) as encoder to get user embedding and attribute embedding, and regards inner product layer as decoder to generate user attributes from the two embedding with missing ones inferred.
HAGT~\cite{HGAT} uses Graph Attention Networks (GAT) as encoder to get user embedding and uses softmax function to get the final results. We argue that a simple combination of the GNN and VAE will generate representations containing too much noise. What's more, their simple decoders lacks adequate inference ability to achieve satisfactory results on attribute inference.

Recently, inspired by GANs, the generative model achieved state-of-art performance in computer vision~\cite{GAN,progressiveGAN,largeGAN}, there are works trying to improve the performance of VAEs through adversarial training~\cite{ALA,AVB,cvae-gan} on images. A few researchers starts to use this insight on non-Euclidean data.
ARGA~\cite{ARVGA} adapts adversarial training approach on VAE by forcing the embedding learned form encoder to match a prior distribution which is Gaussian distribution. AVAE~\cite{AVAE} proposes a framework of VAE generating two kinds of embedding from the input data and trains a discriminator the discern the two, so that the robustness of VAE can be improved. We argue that our model is different from AVAE in that the two representations are generated for distinct purposes and the adversarial training strategy is totally different. The dual representations are learned by leveraging the characteristics of user attribute data in social networks, and adversarial training is leveraged to overcome over-smoothing in GNNs.

Meanwhile, some researchers also tried to improve VAEs performance on some specific problems with adversarial training. For example, VAEGAN~\cite{VAEGAN} introduced an auxiliary discriminator to VAE to better approximate the data posterior, so that the model can get better performance on collaborative filtering. AGAE
~\cite{AGC} developed an adversarial regularizer to train the encoder with an adaptive partition-dependent prior in order to improve VAE's  performance on clustering. All in all, those methods are either general models~\cite{ARVGA,AVAE} or developed for other applications~\cite{VAEGAN,AGC}. Most of them are usually designed to regulate latent representations to a certain distribution, while the data from social network can hardly be fitted by a simple distribute, as a result, these existing methods fail in meeting the need of attribute inference and achieving satisfactory results. 

\section{Problem Formulation}

In this section, we will introduce some formal notations and formulate the problem of attribute inference. 

\subsection{Attributed Network}

Given a social network $\mathcal{S}$, we represent its set of users as $\mathcal{V}$, their links as $\mathcal{E}$, and their attributes as $\mathcal{A}$. In our settings, users have multiple attribute types\footnote{In this paper, we use attribute type and attribute interchangeably.}, e.g., gender, employer, college. Each attribute type can have multiple attribute labels, for example, gender can have attribute label male, female, and so on. Therefore, $\mathcal{S}$ can be represented as an attributed network $\mathcal{G} = (\mathcal{V},\mathcal{E},\mathcal{A})$. 
$\forall \textbf{a}_i \in \mathcal{A}$ is the attribute vector for $v_i$ which records user labels of $L$ attributes. 
$a_{ij}$ is a nonzero integer indicates which attribute label of attribute type $j$ user $v_i$ owns. For example, if $j = 0$ indicates the attribute gender, $a_{ij} = 1$ indicates the gender of user $v_i$ is male, $a_{ij} = 2$ indicates her gender is female, while $a_{ij} = 0$ means $v_i$'s gender is unknown (attribute missing). For user with no missing attributes (no $0$ in $a_{i}$), their attributes are complete, and we represent them as $\mathcal{V}^L$, and the rest users, whose attribute data is incomplete, are represented as $\mathcal{V}^U$ . 

For convenience, we introduce adjacency matrix $\textbf{A}$ and attribute 
matrix $\textbf{X}$ to represent graph $\mathcal{G}$. For adjacent matrix $\textbf{A} \in \mathbb{R}^{N\times N}$, where $N$ is the number of users, $\textbf{A}_{ij} = 1$ if there's a link between $v_i$ and $v_j$, otherwise $\textbf{A}_{ij} = 0$. For each attribute type, we convert $\textbf{a}_{:,j}$ (each colomn of $\mathcal{A}$) to one-hot encoding vectors (zero vectors for $a_{ij} = 0$) and concatenate one-hot encoding vectors of different attributes as users' feature vectors and produce $\textbf{X} \in \mathbb{R}^{N\times F}$, where $F$ is the total number of attribute labels. $\textbf{X}_{ij} = 1$ if $v_i$ owns label $j$, otherwise $\textbf{X}_{ij} = 0$.

\subsection{Attribute Inference}

Given an attributed network $\mathcal{G}$, we regard attribute inference as a semi-supervised problem which aims at inferring missing attribute values for users in $\mathcal{V}^U$ using observed attribute data and large-scale unsupervised information in social connections. The formal format of attribute inference is as follows, 
\begin{equation}
(\textbf{A},\textbf{X})\stackrel{\Xi}{\longrightarrow} \hat{\textbf{X}}
\end{equation}
where $\Xi$ is the inference model and $\hat{\textbf{X}}\in \mathbb{R}^{N\times F}$ is the output attribute matrix after inference. $\hat{\textbf{X}}$ is a real value matrix indicating the probability of user owning each attribute label. 

\section{Methodology}

In this section, we first give a description of our model's general framework. Then we introduce each part of the model in detail and finally, illustrate how to optimize our model.

\begin{figure*}[bp]
	\centering
		\includegraphics[scale=.6]{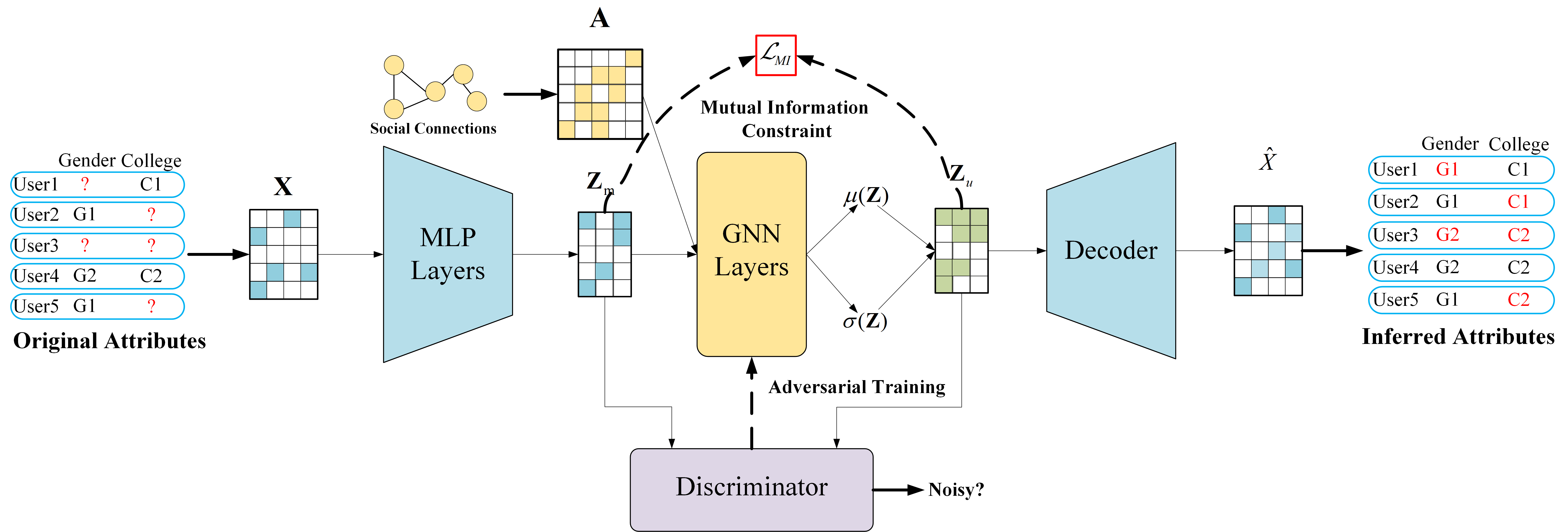}
\caption{\textbf{Framework of Infer-AVAE.}}
\label{fig:frame}       
\end{figure*}

\subsection{Framework}
Under a VAE framework for attribute inference, the encoder of VAE takes observed attributes $\textbf{X}$ as input and produces latent representations by reparameterization trick. Then, the decoder uses the learned representations and generates reconstructed user attributes $\hat{\textbf{X}}$, which hopefully will complete the missing attributes in $\textbf{X}$.
However, we find that VAE can easily suffer from over-fitting and over-smoothing in attribute inference. As shown in Figure \ref{fig:mlp_gcn}, using MLPs as the encoder can match the training data well but the representations contains little useful information for inference, which leads to low accuracy in test set. While implementing the encoder with GNNs results in upsurge of performance due to aggregating promisingly useful information from neighbourhood in the learned representations. But recent studies found that GNNs have the issue of over-smoothing that they generate indistinguishable representations~\cite{deeper}. Over-smoothing hinders GNNs from improving models performance in attribute inference. 

To solve the challenges mentioned above, we propose our model called Infer-AVAE to infer missing attribute values in the input attribute matrix based on adversarial variational autoencoders, which is illustrated in Figure \ref{fig:frame}. Our model is consisted of 4 components: MLP layers, GNN layers, decoder, and a discriminator.
Specifically, to overcome the over-smoothing problem of GNNs, our model unifies VAE with adversarial network. Dual latent representations are generated from MLP layers and GNNs layers. MLP layers encode only the observed attributes into mid latent representations. GNN layers converges these mid latent representations according to social connections and produce user latent representation containing extra information. Then the adversarial network is employed to leverage the differences between the two representations and GNN layers is trained to aggregate less irrelevant information and learn more robust representations via adversarial training. 
After that, to improve the inference ability of decoder and alleviate over-fitting, we specifically design an extra regularization term in loss function. The regularizer measures the mutual information between dual latent representations and encourages the decoder to make use of extra information in the learned representations for attribute inference. Next, we will elaborate on the details of our model.

\subsection{Generating Dual Latent Representations}
In this section, we will describe how to generate dual latent representation for each user.
\subsubsection{Embedding Observed Attributes}
First, MLP layers are employed to embed users' observed attributes, which are represented by attribute matrix $\textbf{X}$ here, into low-dimensional vector space and produce mid latent representations $\textbf{Z}_m$,
\begin{equation}
q_{\phi_{1}}(\textbf{Z}_m|\textbf{X}) = f_{MLP}(\textbf{Z}_m;\textbf{X},\phi_{1})
\end{equation}
where $f_{MLP}$ denotes the MLP layers which is consisted of multi-layer perceptrons~\cite{repara}, while $\phi_{1}$ denotes parameters of MLP.
It is obvious that $\textbf{Z}_m$ only contains observed attribute data, while for users in $\mathcal{V}^U$ whose input attribute data is incomplete, the learned $\textbf{Z}_m^{U}$ is not useful for inferring missing attributes. To handle this limitation, we further leverage social connections to learn representations containing extra information.
\subsubsection{Converging Information Through Graph Structure}
Homophily is a common phenomenon in social networks, which means users tend to have connections with those sharing similar attributes~\cite{homophily,homo}. Inspired by homophily, we adopt graph neural networks to merge $\textbf{Z}_m$ according to network structure and have user latent representation $\textbf{Z}_u$ as output,

\begin{equation}
q_{\phi_{2}}(\textbf{Z}_u|\textbf{Z}_m,\textbf{A}) = f_{GNN}(\textbf{Z}_u;\textbf{Z}_m,\textbf{A},\phi_{2})
\end{equation}
where $f_{GNN}$ denotes GNN layers which in this paper is 2-layer GCNs~\cite{vae-gcn}, while $\phi_{2}$ denotes parameters of GNN layers. The user latent representation $\textbf{Z}_u$ can be learned by,
\begin{center}
\begin{equation}
[\mathbf{\mu}(\textbf{Z}_u),\mathbf{\sigma}(\textbf{Z}_u)]  = \widetilde{\textbf{A}}ReLU(\widetilde{\textbf{A}}\textbf{Z}_m\textbf{W}^{(0)})\textbf{W}^{(1)}
\end{equation}
\begin{equation}
\textbf{Z}_u  = \mathbf{\mu}(\textbf{Z}_u) + \mathbf{\sigma}(\textbf{Z}_u) * \epsilon
\end{equation}
\end{center}
where $\mathbf{\mu}(\textbf{Z}_u)$ and $\mathbf{\sigma}(\textbf{Z}_u)$ are the means and variances of the learned Gaussian embedding, $\epsilon \in \mathcal{N}(0, \bf{I})$ is Gaussian noise variable, $ReLU(\cdot) = max(0, \cdot)$ is the non-linear activation function, $\widetilde{\textbf{A}} = \textbf{D}^{-1/2}
\textbf{A}\textbf{D}^{-1/2}$ is the symmetrically normalized adjacency matrix with 
$\textbf{D}_{ii} 
=\sum_{j}\textbf{A}_{ij}$
being $\mathcal{G}$’s degree matrix, and $\phi_{2}=[\textbf{W}^{(0)},\textbf{W}^{(1)}]$ are trainable weights for GNN layers, respectively.

In this way, the learned $\textbf{Z}_u$ for each user contains not only their own attribute data in the observations, and also abundant extra information from their neighborhood which is promising to be useful for attribute inference due to homophily of users in social network.

\subsection{Improving Robustness of Representations by Adversarial Network}
After incorporating encoder with GNN layers, here comes another problem that GNNs easily suffer from over-smoothing and generate indistinguishable representations.
Over-smoothing deteriorates GNNs' performance in attribute inference severely for inference accuracy highly depends on the expressiveness of user's representations. 
One reason that over-smoothing occurs is that GNNs over-mixes information and noise, the messages gotten from neighbourhood may be useless even harmful~\cite{chen2020measuring}. 
Once $\textbf{Z}_u$ contains too much irrelevant or noise information, the attribute data decoded from it will hardly match users’ real attributes. In this paper, we address this issue by training an adversarial network leveraging the differences between dual latent representations. 

We notice that for users who are in $\mathcal{V}^L$ with complete attribute information in the observed data, the learned mid latent representation $\textbf{Z}_m^{L}$ contains enough information, while the learned user latent representation $\textbf{Z}_u^{L}$ after convolution contains redundant information which, on the contrary, may be noisy. Consequently, we develop an adversarial network to distinguish $\textbf{Z}_m^{L}$ (positive) from $\textbf{Z}_u$ (negative) so that it learns to judge whether there is noise in representations. The adversarial network $D$ is built on two layers of standard MLP where the output layer only has one dimension with a sigmoid function. The objective of the discriminator is as follows,

\begin{equation}
\resizebox{.88\linewidth}{!}{$
    \displaystyle
\mathcal{L}_{D} = -\mathbb{E}_{\textbf{Z}\sim p(\textbf{Z}_{m}^{L})}\log D(\textbf{Z})-\mathbb{E}_{\textbf{Z}\sim p(\textbf{Z}_{u})}\log (1-D(\textbf{Z}))
$}
\end{equation}

After training $D$, we use it to do adversarial training on GNN layers which act as the generator in GAN~\cite{GAN}. In adversarial training, GNN layers aims at generating $\textbf{Z}_u$ and cheating $D$ to regard $\textbf{Z}_u$ as $\textbf{Z}_m^{L}$
The equation for training GNN layers with $D$ can be written as,

\begin{equation}
\mathcal{L}_{GNN} = -\mathbb{E}_{\textbf{Z}\sim p(\textbf{Z}_{u})}\log D(\textbf{Z})
\end{equation}

In orther words, $D$ and GNN layers play the minimax game with value function $V (GNN, D)$:
\begin{equation}
\resizebox{.88\linewidth}{!}{$
    \displaystyle
\mathop{\min}_{f_{GNN}}\mathop{\max}_{D} \mathbb{E}_{\textbf{Z}\sim p(\textbf{Z}_{m}^{L})}\log D(\textbf{Z})-\mathbb{E}_{\textbf{Z}\sim p(\textbf{Z}_{m})}\log (1-D(f_{GNN}(\textbf{Z})))
$}
\end{equation}

After iterative learning, GNN layers will generate robust representations that $D$ is hard to distinguish from the ones generated from MLP layers. In this way, the parameters of GNN layers will be optimized so that the learn user latent representations contain as little noise as possible.
\subsection{Mutual Information Constraint}

The decoder in vanilla VAEs is used to decode information from the learned latent representation and reconstruct observations while for attribute inference, we use the decode to generate $\hat{\textbf{X}}$ with the missing attribute values inferred using user latent representations $\textbf{Z}_u$,

\begin{equation}
p_{\theta_{de}}(\hat{\textbf{X}}|\textbf{Z}_u) = f_{DEC}(\hat{\textbf{X}};\textbf{Z}_u,\theta_{de})
\end{equation}
where $f_{DEC}$ denotes the decoder, which in this paper is MLP with parameters $\theta_{de}$, $\hat{\textbf{X}}$ denotes the reconstructed attribute matrix. The encoder and decoder in VAE will be trained as a whole by optimizing the loss function called evidence lower bound (ELBO) using the reparameterization trick~\cite{repara} and stochastic gradient descent,
\begin{equation}
\begin{aligned}
\mathcal{L}_{VAE} = -\mathbb{E}_{\textbf{Z}_u\sim q_{\phi_{en}}(\textbf{Z}_u|\textbf{X},\textbf{A})}[\log p_{\theta_{de}}(\hat{\textbf{X}}|\textbf{Z}_u)] \\
+KL(q_{\phi_{en}}(\textbf{Z}_u|\textbf{X},\textbf{A})||p(\textbf{Z}_u))
\end{aligned}
\end{equation}
where $\phi_{en} = [\phi_{1},\phi_{2}]$ denotes parameters of the encoder and GNN layers and $KL(\cdot||\cdot)$ is the Kullback-Leibler (KL) divergence.

From the loss function above, we observe that the main objective of VAE is to reconstruct the input attribute matrix while hardly addresses the problem of inferring the missing part, which doesn't meet the needs of attribute inference and makes VAEs suffer from severe over-fitting. In order to adapt VAE to attribute inference and improve decoder's inference ability, we specifically design an additional term which is mutual information constraint in the loss function for training decoder.

As mentioned above, the learned mid latent representation $\textbf{Z}_m$ only encodes the information from observations, which means if send $\textbf{Z}_m$ to decoder directly, the output $\hat{\textbf{X}}_m$ can match the input attribute matrix well but fail in inferring the missing one,
\begin{equation}
p_{\theta_{de}}(\hat{\textbf{X}}_m|\textbf{Z}_m) = f_{DEC}(\hat{\textbf{X}}_m;\textbf{Z}_m,\theta_{de})
\end{equation}
where $f_{DEC}$ denotes decoder the same as Eq.(7), while $\hat{\textbf{X}}_m$ denotes the reconstructed attribute data directly from $\textbf{Z}_m$.

Now, we have two pairs of reconstructed attribute matrix: $(\hat{\textbf{X}}_m^{L}$,$\hat{\textbf{X}}^{L})$, which belongs to users in $\mathcal{V}^L$ with complete attributes in observations; $(\hat{\textbf{X}}_m^{U}$,$\hat{\textbf{X}}^{U})$, which belongs to users with unknown attributes in the input data $\mathcal{V}^U$. For an ideal decoder, we want the first pair to be close to each other because $\hat{\textbf{X}}_m^{L}$ has the exact user information and $\hat{\textbf{X}}^{L}$ should just be the same; while the later pair should be different from each other because $\hat{\textbf{X}}_m^{U}$ contains little information and $\hat{\textbf{X}}^{U}$ should decode as much extra information from user latent representations as possible. To implement such instinct, we introduce an extra term in loss function called mutual information constraint to train the decoder as a discriminator and use mutual information to evaluate the difference within each pair,
\begin{equation}
\mathcal{L}_{MI} = -MI(\hat{\textbf{X}}_m^{L},\hat{\textbf{X}}^{L}) + MI(\hat{\textbf{X}}_m^{U},\hat{\textbf{X}}^{U})
\end{equation}
The mutual information constraint aims at maximizing mutual information between $(\hat{\textbf{X}}_m^{L}$,$\hat{\textbf{X}}^{L})$ while minimizing mutual information between $(\hat{\textbf{X}}_m^{U}$,$\hat{\textbf{X}}^{U})$.
Following the previous researches, we get mutual information here by calculating the lower bounds on MI, which is InfoNCE~\cite{InfoNCE},
\begin{equation}
MI(X,Y)\ge\mathbb{E}[\frac{1}{K}\sum^K_{i=1}log\frac{e^{f(x_i,y_i)}}{\frac{1}{K}\sum^K_{j=1}e^{f(x_i,y_j)}}]
\end{equation}
where the expectation is over $K$ independent samples, for the first pair ($\hat{\textbf{X}}_m^{L}$,$\hat{\textbf{X}}^{L}$), $K = |\mathcal{V}^L|$; for the second pair ($\hat{\textbf{X}}_m^{U}$,$\hat{\textbf{X}}^{U}$), $K = |\mathcal{V}^U|$. $f(x, y)$ is critic function and we choose the inner product critic~\cite{innerproduct} $f(x, y)=x^{T}y$.

\begin{algorithm}[tb]
\caption{Infer-AVAE}
\label{alg:algorithm}
\textbf{Input}: adjacency matrix $\textbf{A}$, attribute matrix $\textbf{X}$, iterations $\tau$, hyper-parameter $\beta$ and $\lambda$. \\
\textbf{Parameter}: network parameters $\phi_{en} = [\phi_{1},\phi_{2}]$, $\theta_{de}$, $\theta_{D}$ \\
\textbf{Output}: reconstructed attribute matrix $\hat{\textbf{X}}$ \\
\begin{algorithmic}[1] 
\STATE Initialize network parameters.
\FOR{$i = 1$ to $\tau$}
\STATE 	Generate dual latent representations $\textbf{Z}_{m}$ and $\textbf{Z}_{u}$; 
\STATE  Generate $\hat{\textbf{X}}$ and $\hat{\textbf{X}}_{m}$ from $\textbf{Z}_{u}$ and $\textbf{Z}_{m}$; 
\STATE  Update $\phi_{en}, \theta_{de}$ by minimizing $\mathcal{L}_{VAE} + \lambda\mathcal{L}_{MI}$ in Eq.(14);
\STATE  Update $\theta_{de}$ by minimizing $\mathcal{L}_{G}$ in Eq. (12);
\STATE  Sample $\textbf{Z}_m^{L}, \textbf{Z}_u^{L}$ from $\textbf{Z}_{m}, \textbf{Z}_{u}$;
\STATE  Update $\theta_{D}$ by minimizing $\mathcal{L}_{D}$ in Eq.(6);
\STATE  Update $\phi_{2}$ by minimizing Eq.(7);
\ENDFOR
\STATE Generate $\hat{\textbf{X}}$ from $\textbf{Z}_{u}$
\STATE \textbf{return} $\hat{\textbf{X}}$
\end{algorithmic}
\end{algorithm}

\subsection{Model Optimization}

The final loss function of our model is as follows,
\begin{equation}
\mathcal{L} = \mathcal{L}_{VAE} + \beta\mathcal{L}_{GAN} + \lambda\mathcal{L}_{MI}
\end{equation}
where $\mathcal{L}_{GAN} = \mathcal{L}_{D} + \mathcal{L}_{GnN}$ and all terms are presented in former equations. $\beta$ and $\lambda$ are hyper-parameters. All the hyper-parameters will be further discussed in Section 5.4.

To be specific, Infer-AVAE will be optimized according to Algorithm 1. Given attribute matrix $\textbf{X}$ and adjacent matrix $\textbf{A}$, in step 3, we generate dual latent representations $\textbf{Z}_m$ and $\textbf{Z}_u$ from the encoder. Then, in step 4, $\textbf{Z}_m$ and $\textbf{Z}_u$ will be sent to decoder respectively and decoder produces $\textbf{X}_m$ and $\hat{\textbf{X}}$ correspondingly. The loss of VAE and mutual information constraint will be minimized and back-propagated at the same time. Then comes the adversarial training. In step 7, $\textbf{Z}_m^{L}$ and $\textbf{Z}_u$ is sampled and used to train the adversarial network $D$ in step 8. Parameters of GNN layers will be further updated by the output of adversarial network in step 9.
After multiple iterations of training  when the whole model finally converged, we will return the reconstructed attribute matrix $\hat{\textbf{X}}$ as the result of attribute inference.

\paragraph{Time Complexity.}
The time cost per iteration is comprised of three parts: i) Optimizing over encoder and decoder as a whole gives $\mathcal{O}(|\mathcal{V}|FD+|\mathcal{E}|D^3+|\mathcal{V}|FHD)$ where $F$, $D$ and $H$ is the dimension of $X$, $Z_m$ and hidden layer. ii) Optimizing over adversarial network and adversarial training on GNN layers is $\mathcal{O}(|\mathcal{V}^L|FHD + \mathcal{E}D^3)$. iii) The time cost of additional regularizer on decoder is $\mathcal{O}(|\mathcal{V}|FHD)$. With all hyperparameters fixed, the overall time complexity per iteration can be regarded as $\mathcal{O}(|\mathcal{E}|+|\mathcal{V}|)$ which is on par with the baseline methods such as Graph Attention Networks~\cite{GAT}.

\section{Experiments}
We provide experimental results to demonstrate the effectiveness of our proposed Infer-AVAE on real-world social network datasets. The experiments are designed to evaluate Infer-AVAE from three aspects: (1) Accuracy on attribute inference (2) Sensitiveness to different levels of label sparsity (3) Effectiveness of Infer-AVAE's each component 
\subsection{Experimental Settings}
\subsubsection{Datasets}
To evaluate our proposed method in attribute inference, we chose three public social network datasets: Facebook100-Carnegie49 (fb-CMU), Facebook100-Carnegie49 (fb-
Harvard)~\cite{FB100}, 
Facebook dataset (fb-SNAP) from SNAP 
social network datasets~\cite{snap}. 
What's more,
we collect data from Facebook (fb-Ours) to further evaluate proposed 
model in inferring different attributes. 
Statistics of the four datasets are shown in Table \ref{table:dataset}. These social network datasets have multiple attribute types and corresponding attribute labels. Here, attribute types means attribute types like gender, education, employer, and soon. Labels means attribute labels under attribute types, e.g.,label "male" and "female" of attribute gender.

In fb-CMU dataset, users have 6 attribute types including student/faculty status (6 labels), gender (2 labels), major (41 labels), second major/minor (42 labels), dorm/house (51 labels) and year (18 labels).
In fb-Harvard dataset, users have 6 attributes including student/faculty status (6 labels), gender (2 labels), major (59 labels), second major/minor (59 labels), dorm/house (42 labels) and year (53 labels).
In fb-SNAP dataset, users have 5 attributes including gender (2 labels), education (34 labels), location (75 labels), year of birth (40 labels), and hometown (98 labels).
In fb-Ours dataset, users have 4 attributes including gender (2 labels), education (279 labels), and hometown (218 labels).

\renewcommand{\arraystretch}{1.5} 
\begin{table}[b]  
  \centering  
    \begin{tabular}{lrrrr}  
    \toprule  
    \textbf{Datasets}&\textbf{\#Nodes}&\textbf{\#Edges}&\textbf{\#Attributes}&\textbf{\#Labels}\cr  
    \midrule  
    fb-CMU&6637&497775&6&160\cr
    fb-Harvard&15126&1648734&6&221\cr
    fb-SNAP&4035&173406&5&249\cr 
    fb-Ours&1721&18376&4&498\cr 
    \bottomrule  
    \end{tabular}  
    \caption{\textbf{Statistics of real-world social network datasets.} Attributes means attribute types like gender, education, employer, and so on. Labels means attribute labels under attribute types, e.g., label "male" and "female" of attribute gender. }  
    \label{table:dataset}  
\end{table}

\subsubsection{Baselines}
We compare our proposed model with the following methods: vanilla VAE, 2 state-of-the-art attribute inference models (CAN and BLA), and 2 graph neural networks showing the best performance on graph representation learning (GCN and GAT). Note that GCN and GAT cannot directly used for attribute inference. In our experiments, we implement them under the framework of VAE, implementation details are shown in Section 5.1.4.

\paragraph{VAE} Variational Auto-Encoder embeds input data into low dimensional representations that are sampled from Gaussian priors and reconstructs data through decoder~\cite{repara}. 

\paragraph{CAN} Co-embedding Attributed Networks learns the low-dimensional representations of nodes and attributes in the same semantic space basing on VAEs~\cite{CAN}. 

\paragraph{BLA} Bi-directional joint inference for user Links and Attributes utilizes the mutual reinforcement between links and attributes and improves label propagation with link prediction~\cite{BLA}.

\paragraph{GCN} Graph Convolution Network is a semi-supervised learning 
algorithm on 
data of graph structure. 
It is widely used for node classification~\cite{semi-gcn}. 

\paragraph{GAT} Graph Attention Network is one of state-of-the-art graph neural network models. By learning different attention coefficients to neighbors, it can acquire a better representation of nodes~\cite{GAT}. 

\begin{table*}[tp]  
  \centering  
  \begin{threeparttable}  
    \begin{tabular}{ccccccccc}  
    \toprule  
    \multirow{2}{*}{Method}&  
    \multicolumn{2}{c}{fb-CMU}&\multicolumn{2}{c}{fb-Harvard}&\multicolumn{2}{c}{fb-SNAP}&\multicolumn{2}{c}{fb-Ours}\cr
    \cmidrule(lr){2-3} \cmidrule(lr){4-5} \cmidrule(lr){6-7} \cmidrule(lr){8-9}
    &$Accuracy$&$Macro$-$F_1$&$Accuracy$&$Macro$-$F_1$&$Accuracy$&$Macro$-$F_1$&$Accuracy$&$Macro$-$F_1$\cr  
    \midrule  
    VAE&0.3812&0.077&0.2531&0.0319&0.2396&0.015&0.3014&0.1372\cr
    CAN&0.4293&0.4316&0.4751&0.4986&0.4455&0.4374&0.4254&0.3659\cr  
    BLA&0.5795&0.5486&0.5353&0.5414&0.4180&0.4348&0.4110&0.4244\cr  
    GAT&0.6023&0.6232&0.5323&0.5836&0.4398&0.4721&0.5307&0.4377\cr  
    GCN&0.6099&0.6204&0.5693&0.5678&0.4171&0.4619& 0.5043&0.4093\cr
    \midrule 
    Infer-AVAE w/o A&0.6034&0.6107&0.5059&0.5149&0.4540&0.4818&0.4881&0.3988\cr
    Infer-AVAE w/o M&0.6216&0.6250&0.6165&0.6153&0.4563&0.4899&0.4836&0.3799\cr
    Infer-AVAE&{\bf 0.6542}&{\bf0.6666}&{\bf0.6431}&{\bf0.6413}&{\bf0.4737}&{\bf0.4975}&{\bf 0.5613}&{\bf0.4644}\cr
    \bottomrule  
    \end{tabular}  
    \caption{\textbf{Attribute Inference performance comparison by $Accuracy$ and $Macro$-$F_1$.} Best results are shown in \textbf{bold}. w/o indicates Infer-AVAE without some component for ablation analysis.}  
  \label{tab:performance_comparison}  
 \end{threeparttable}  
\end{table*}  

\subsubsection{Evaluation Metrics}
In the experiments, we choose two metrics $Accuracy$ and $Macro-F1$, which are widely used in attribute inference and user profiling problems~\cite{eval}, to evaluate the performance of our model. In the testing process, if the  $j$th attribute of user $i$'s true label of is $k$ and, according to $\hat{\textbf{X}}$, the predicted value of user $i$'s $k$ label is the largest in the labels of user $i$'s attribute $j$, then we regard user $i$'s attribute $j$ as correctly inferred, which means label $k$ of user $i$ is true positive and the other labels of user $i$'s attribute $j$ is true negative. For example, if user $i$'s true label of attribute "gender" (attribute $j$) is "female" (label $k$), when "female" label is larger than her other gender labels in $\hat{\textbf{X}_{i}}$, which is "male" label, then "female" label is true positive for user $i$. The two evaluation metrics are calculated as follows~\cite{evalequation}.

\begin{equation}
Accuracy = \frac{TP+TN}{TP+FP+TN+FN}
\end{equation}
where $TP$ is the number of true positives (correctly inferred
label belongs to the user), $TN$ is the number of true negatives (correctly identified
labels not belongs to the user), FP is the number of false positives (label mistakenly
classified as belonging to the user), and FN is the number of false negatives (label mistakenly classified as not belonging to the user).

\begin{equation}
Macro-F1 = \frac{1}{L}\sum_{i=1}^{L}\frac{2*P_{i}*R_{i}}{P_{i}*R_{i}}
\end{equation}
where $L$ is the number of attributes, $P_{i}$ is the precision of attribute $i$, and $R_{i}$ is the recall of attribute $i$.

We conduct each experiment 20 times and report the mean values as the final scores.

\begin{figure*}[tp]
\centering    
\subfigure[fb-CMU] 
{
	\label{fig::his_CMU}
	\includegraphics[scale=.5]{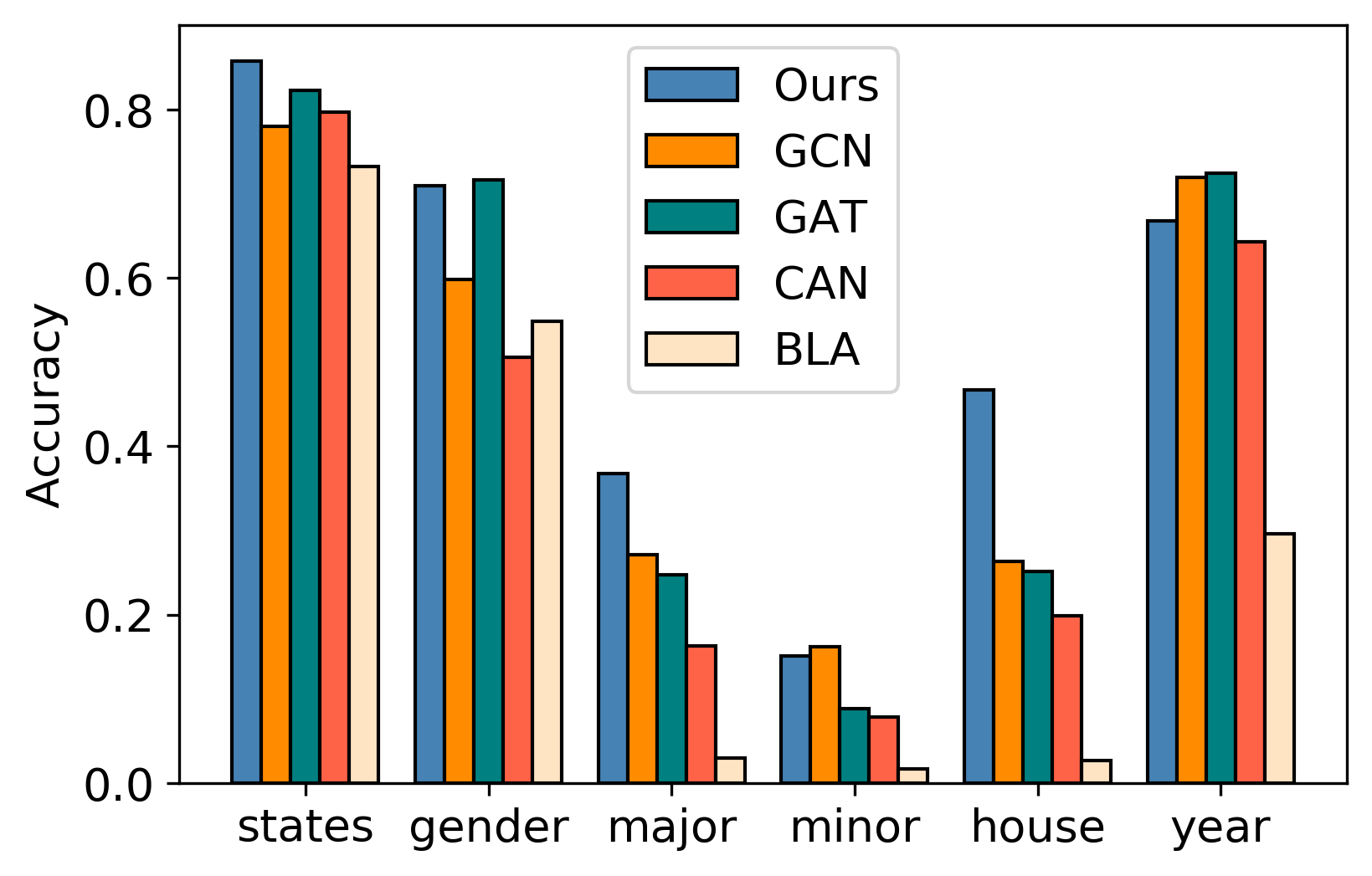} 
}
\subfigure[fb-Harvard] 
{
	\label{fig::his_Har}
	\includegraphics[scale=.5]{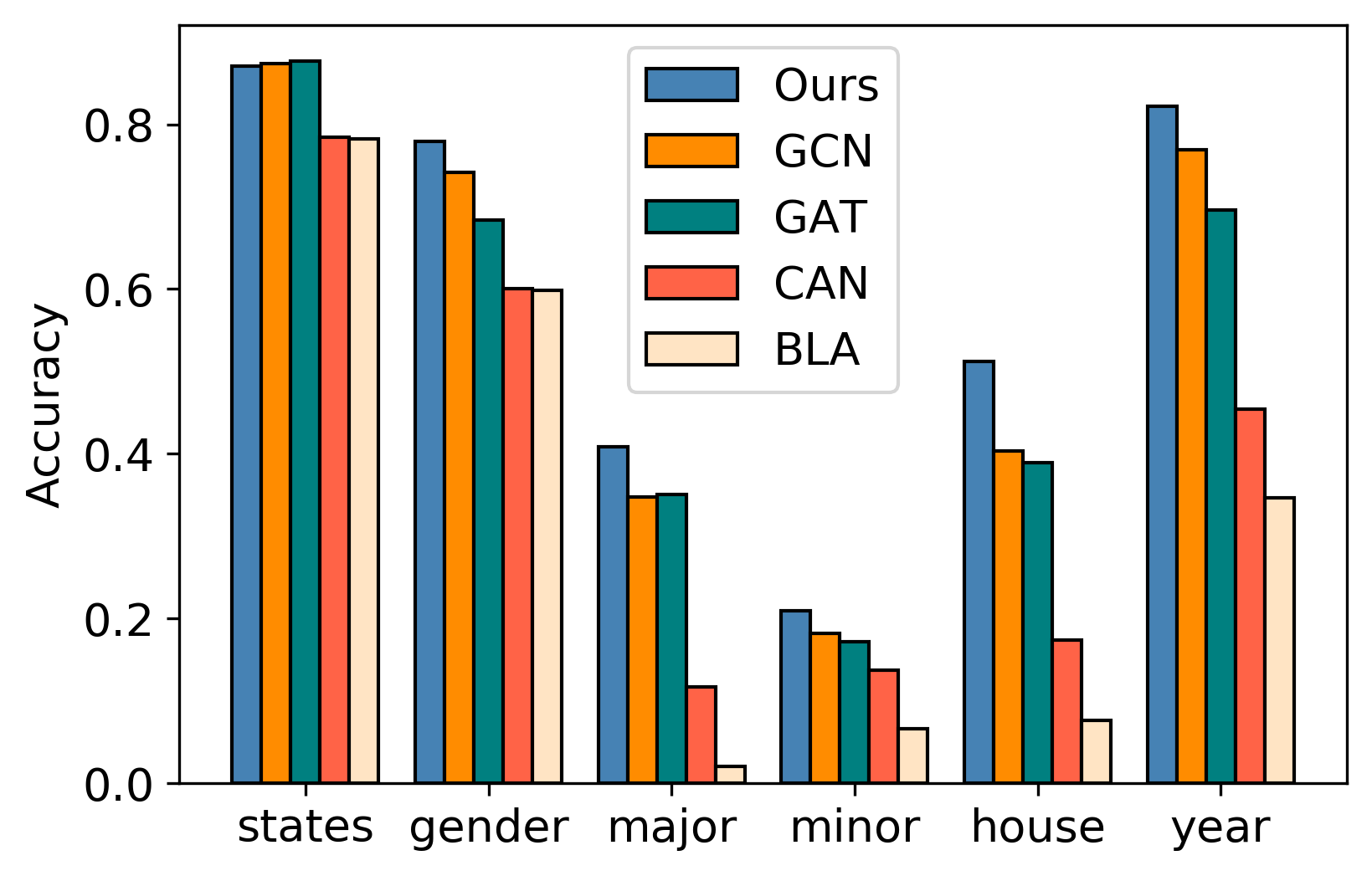} 
}

\label{fig:his}  
\caption*{\textbf{Accuracy of each attribute in datasets fb-CMU and fb-Harvard.}
} 
\end{figure*}

\subsubsection{Implementation Details}
For all methods, we randomly split existing labeled user attributes into training set, validation set and test set with the ratio 80:10:10 following previous works~\cite{CAN}. The edges between users are all kept in the training set.
During the training stage, we use the latent representations of all
relevant users, but only the labels assigned to the users in the training set. In the validation and testing stages, we use the labels of users in the validation and test set to evaluate the methods.

For baselines, we implemented both vanilla VAE's encoder and decoder with 2 layers of MLP. For GCN and GAT, we implemented them under the framework of VAE. We used them separately as the encoder of VAE, and the decoder is implemented with 2 layers of MLP. For We implemented all the baselines based on the codes released by the authors. The hyperparameters of all baselines are tuned to be optimal.

For Infer-AVAE, the dimension of MLP layer is 64, GNN layers consist of 2 layers of 64-dimensional GCN. The decoder is a 128-dimensional hidden layer and output layer, both of which are MLP. The learning rate for encoder and decoder is 0.01.
Adversarial network is built with two hidden layers(16-
dimensional, 4-dimensional MLP respectively) and there is a softmax layer to obtain the categorical probability before the output layer. Its learning rate is set as 0.001. We train our model for 500 iterations by Adam optimizer. The two hyper-parameters $\lambda$ and $\beta$ are set to 0.2 and 0.3, respectively. The discussion on the influence of hyper-parameters is in Section 5.4.

\subsection{Comparison with  Baselines}

\subsubsection{Overall comparison.}
Table \ref{tab:performance_comparison} shows the results between our proposed model and the baselines mentioned above. Infer-AVAE achieves the best performance on four datasets with both evaluation metrics, which illustrates the effectiveness of our proposed model. Next, we analyze the results from 3 aspects.

\textbf{First, Infer-AVAE outperforms all the baselines in the experiments.} 
In detail, we observe that vanilla VAE suffers from severe overfitting (also shown in Figure \ref{fig:mlp_gcn}) and achieves low performance while CAN, GAT, and GCN have significant performance improvements after incorporating graph neural networks, which demonstrates the effectiveness of GNNs in learning more expressive latent representations. Infer-AVAE outperforms these GNN-based methods for generating more meaningful representations after filtering out noise information with the help of adversarial network. Moreover, the decoder of our model can make better use of the information in latent representations under the training of mutual information constraint. We notice that even though BLA and CAN achieved the state-of-art results under the evaluation metrics like AUC (Area Under Curve) and Precision@K \cite{BLA, CAN}, they failed in showing satisfying results in our experiment settings where we aim to infer user's specific label under each attribute. Such inference results are more meaningful and the completed user attributes are more useful for  downstream tasks like user profiling and personalized recommendation. 

\textbf{Second, Infer-AVAE achieves the best performance across 4 datasets.}
Take $Accuracy$ for example, our model achieves $7.2\%$, $13.0\%$, $1.9\%$, $5.8\%$ improvement over the most competitive method on the four datasets respectively. We observe that the performance of attribute inference models is highly influenced by the characteristics of datasets. Both Infer-AVAE and the baselines achieve relatively higher accuracy and Macro-F1 scores in datasets fb-CMU and fb-Harvard than in fb-SNAP and fb-Ours. That may be due to the reason that fb-CMU and fb-Harvard were collected within colleges where users are densely connected and the phenomenon of homophily is more obvious. While the data of fb-SNAP and fb-Ours was collected from open social networks where users may build relationship with each other due to various reasons, which is hard to infer their attributes according to social connections. In baseline methods, we find that GCN shows the most competitive performance in fb-CMU and fb-Harvard datasets, which indicates its advantage of leveraging task-specific graph structure. When it comes to fb-SNAP and fb-Ours datasets, GAT shows the ability to handle irrelevant linkages due to self-attention mechanism. But facing different datasets, Infer-AVAE is still able to achieve the best results, which illustrates the robustness of our proposed model.

\textbf{Third, Infer-AVAE achieves stable performance in different attribute types.}
As shown in Figure \ref{fig:his}, we explore our model's effectiveness on each attribute type in datasets fb-CMU and fb-Harvard, both of which have 6 attributes. According to the histograms, our model not only achieved better overall performance but also has higher accuracy in most of attributes. In a nutshell, all the methods show similar trends that they achieved high accuracy in attributes with fewer labels (states, gender, and year) but had difficulty in inferring attributes with more labels (major, minor, and house). The phenomenon is caused by sparser user distribution in attributes with fewer labels and there are also fewer training (user, label) samples for models to learn. We observe that in attribute house that other baselines performed badly but our model is able to improve the accuracy to a large degree. This result may be caused by denoising the learned representation and decoder's improved inference ability. After filtration, even though the input data of the attribute ``house'' is inadequate for inference, the representation can capture the correlation between ``house'' and other attributes so that it contains more information for decoder to generate and make more accurate inference. The decoder is able to make use of such auxiliary information by evaluating the mutual information between dual latent representations.

\begin{figure}[tp]
\centering    
\subfigure[fb-CMU] 
{
	\label{fig::CMU}
	\includegraphics[scale=.25]{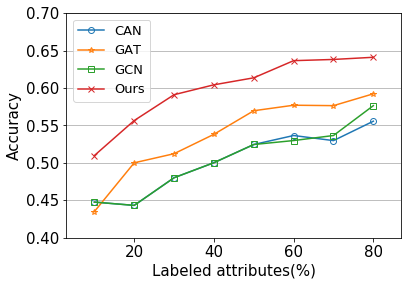}
}
\subfigure[fb-Harvard] 
{
	\label{fig::HAR}
	\includegraphics[scale=.25]{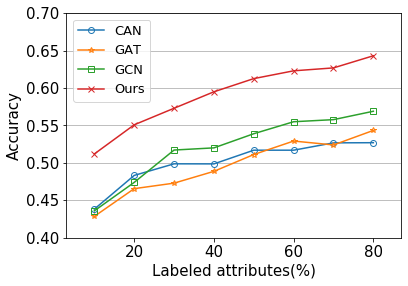}
}
\subfigure[fb-SNAP] 
{
	\label{fig::SNAP}
	\includegraphics[scale=.25]{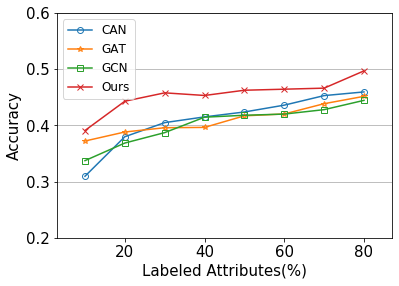}
}
\subfigure[fb-Ours] 
{
	\label{fig::OURS}
	\includegraphics[scale=.25]{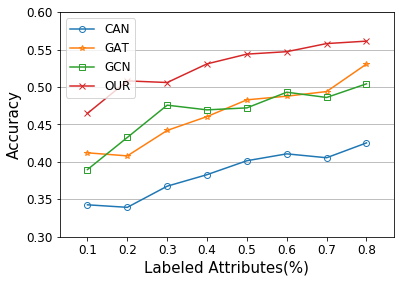}
}
\caption{\textbf{Performance under different degree of label sparsity on datasets fb-CMU, fb-Harvard, fb-SNAP, and fb-Ours datasets.} Label "Ours" indicates our proposed model Infer-AVAE.} 
\label{fig:sparsity}  
\end{figure}

\subsubsection{Performance under different settings of label sparsity.}
Label sparsity is a factor we particularly care about in real world social network. We may face different sparsity of user attributes, the ideal attribute inference model should be robust enough to handle these conditions. In this experiment, we trained Infer-AVAE and baseline methods with different ratios of labeled attributes and compared their performances. Here, we set the test set fixed with 20\% of all labeled attributes while changing the training set containing labeled attributes ranges from 10\% to 80\%. Note that there has already been users missing attribute labels in these datasets, and this experiment further worsened data sparsity.

We report the results on the 4 datasets in Figure \ref{fig:sparsity}. When the ratio of labeled attributes decreases on datasets these datasets, the performance of all methods will drop accordingly, while Infer-AVAE achieves the best results in all conditions across the 4 datasets, which indicates the robustness of our proposed model. The superiority of Infer-AVAE may because when training data is sparse, the decoder is encouraged to make more use of the extra information which GNN layers  converges from the neighborhood. The decoder will not be restricted by the training data which is especially useful when training data is extremely sparse. While other methods will learn little knowledge due to limited input attribute data. When the number of labelled attributes increases, the adversarial network can guide GNN layers to filter out redundant information and generate robust representations, which leads to better performance than the baselines.

\subsection{Ablation Analysis}
\subsubsection{Effectiveness of Adversarial Network.}

Table \ref{tab:performance_comparison} (Infer-\\
AVAE w/o A) shows the results of our model on each dataset without adversarial network and adversarial training. By comparing Infer-AVAE w/o A with Infer-AVAE, we find that the performance will drop a lot using user representation gotten from GNN layers without adversarial training's guidance and is no better than other GNN-based methods. That is reasonable because adversarial training is the key component of Infer-AVAE.
As we have mentioned, the performance of VAE-based models depends heavily on the latent representations. Even though the representations GNN layers generated contain extra information can result in better performance than MLP (vanilla VAE), the embeddings contain too much noise, which hinders the further improvements on accuracy of attribute inference. The adversarial network can help regulate the GNN layers filter out noise information and learn more useful representations, which lays a solid foundation for accurate inference.

\subsubsection{Effectiveness of mutual information constraint.}

Table \ref{tab:performance_comparison} (Infer-AVAE w/o M) shows the results of our model on each dataset without the regularization term, which is the mutual information constraint, in the loss function . Compare Infer-AVAE w/o A with Infer-AVAE, we find that the performance will drop accordingly without extra training.
Only by learning to discriminate the two dual latent representations through evaluating mutual information, can the decoder better leverage the auxiliary information in user latent representations which is converged by GNN layers. After the training of the regulatrization term, the inference ability of the decoder is thus promoted and it can reconstruct data not limited by the input and alleviate overfitting. 

\begin{figure}[tp]
\centering    
\subfigure[$\lambda$] 
{
	\label{fig::sigma}
	\includegraphics[scale=.25]{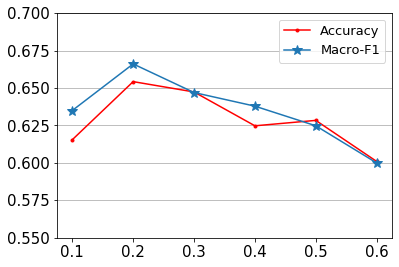} 
}
\subfigure[$\beta$] 
{
	\label{fig::T}
	\includegraphics[scale=.25]{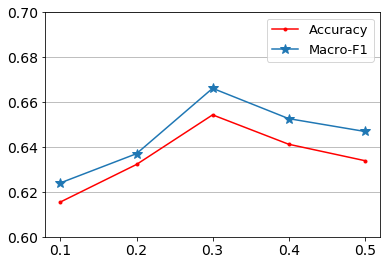} 
}
\caption{\textbf{Analysis of parameter $\lambda$ and $\beta$ in dataset fb-CMU.}} 
\label{fig:parameter}  
\end{figure}

\subsection{Influence of Key Hyperparameters}
We investigate the influence of two key hyperparameters in this section. Specifically, we evaluate how the hyper-parameter $\lambda$ and $\beta$ affect the performance. We conduct this experiment on four datasets and obtain a similar trend. So we only report the results on the fb-CMU dataset for brevity in Figure \ref{fig:parameter}. 
\subsubsection{Influence of $\lambda$}
For the influence of hyperparameter $\lambda$, $\lambda$ indicates the degree of influence the regularizer mutual information constraint $\mathcal{L}_{MI}$ has on the decoder, that is how close $(\hat{\textbf{X}}_m^{L}$,$\hat{\textbf{X}}^{L})$ should be and, more importantly, how different $(\hat{\textbf{X}}_m^{U}$,$\hat{\textbf{X}}^{U})$ should be. If $\lambda=0$, this means the decoder will not learn to distinguish $\textbf{Z}_m$ from $\textbf{Z}_u$. As $\lambda=0$ gradually rises from 0 to 0.2, as shown in Figure \ref{fig::sigma}, the performance improves. This is because decoder's inference ability improved by making use of the difference between the dual latent representations and thus decodes more information from user latent representation $\textbf{Z}_u$ and generate diverse outputs instead of copying the inputs. In this way, Infer-AVAE is less affected by over-fitting. However, the performance decreases when $\lambda$ goes larger than 0.2. The reason is that when $\lambda$ is too large, the training process will not be stable and generate diverse but useless outputs. 
\subsubsection{Influence of $\beta$}
For the influence of hyperparameter $\beta$, $\beta$ indicates influence of adversarial training on GNN layers. The larger $\beta$ is, the more influence adversarial network has on the whole model. As shown in Figure \ref{fig::T}, with proper adversarial training to regulate GNN layers, which is $\beta = 0.3$ in our experiments, it successfully improves the robustness of user latent representation $\textbf{Z}_u$. However when $\beta$ gets larger, the learned $\textbf{Z}_u$ will be over-regularized by $\textbf{Z}_m$. As a result, GNN layers can hardly converge information from neighbourhood leveraging information in social connections.

\section{Conclusions}
In this paper, we present an attribute inference model unifying VAEs with adversarial network called Infer-AVAE. Our proposed model alleviates the problem of over-smoothing and over-fitting in existing VAE-based attribute inference model and achieves better performance. Infer-AVAE first generates dual latent representations by incorporating encoder with MLP layers and GNN layers, and then adversarial network adopted to leverage the two representations for more robust latent representations. Moreover, by adding the mutual information constraint in loss function, the decoder's inference ability is further improved after being trained to leverage auxiliary information in representations. Extensive experiments on real social networks have not only demonstrated Infer-AVAE's significant performance improvements compared to the baseline methods but also its usability under various degrees of label sparsity.

\section{Acknowledgements}
The research presented in this paper is supported in part by the National Key Research and Development Program of China (2018YFB0803501), the National Natural Science Foundation (U1736205, 61833015, U1766215, U1936110,
61902308) 
, the Fundamental Research Funds for the Central Universities (xzy012019036)
,  Foundation of Xi'an Jiaotong University under grant xxj022019016, and Initiative Postdocs Supporting Program BX20190275.

\printcredits


\bibliographystyle{unsrt}

\bibliography{cas-refs}

\begin{thebibliography}{10}

\bibitem{employers}
Alan Mislove, Bimal Viswanath, Krishna~P Gummadi, and Peter Druschel.
\newblock You are who you know: inferring user profiles in online social
  networks.
\newblock In {\em Proceedings of the third ACM international conference on Web
  search and data mining}, pages 251--260. ACM, 2010.

\bibitem{home}
Rui Li, Shengjie Wang, Hongbo Deng, Rui Wang, and Kevin Chen-Chuan Chang.
\newblock Towards social user profiling: unified and discriminative influence
  model for inferring home locations.
\newblock In {\em Proceedings of the 18th ACM SIGKDD international conference
  on Knowledge discovery and data mining}, pages 1023--1031. ACM, 2012.

\bibitem{homophily}
Miller McPherson, Lynn Smith-Lovin, and James~M Cook.
\newblock Birds of a feather: Homophily in social networks.
\newblock {\em Annual review of sociology}, 27(1):415--444, 2001.

\bibitem{homo}
Aravind Sankar, Xinyang Zhang, Adit Krishnan, and Jiawei Han.
\newblock Inf-vae: A variational autoencoder framework to integrate homophily
  and influence in diffusion prediction.
\newblock In {\em Proceedings of the 13th International Conference on Web
  Search and Data Mining}, pages 510--518, 2020.

\bibitem{Ego}
Rui Li, Chi Wang, and Kevin Chen-Chuan Chang.
\newblock User profiling in an ego network: co-profiling attributes and
  relationships.
\newblock In {\em Proceedings of the 23rd international conference on World
  wide web}, pages 819--830. ACM, 2014.

\bibitem{EdgeExplain}
Deepayan Chakrabarti, Stanislav Funiak, Jonathan Chang, and Sofus~A Macskassy.
\newblock Joint inference of multiple label types in large networks.
\newblock {\em arXiv preprint arXiv:1401.7709}, 2014.

\bibitem{Mobi}
Yuxiao Dong, Yang Yang, Jie Tang, Yang Yang, and Nitesh~V Chawla.
\newblock Inferring user demographics and social strategies in mobile social
  networks.
\newblock In {\em Proceedings of the 20th ACM SIGKDD international conference
  on Knowledge discovery and data mining}, pages 15--24. ACM, 2014.

\bibitem{CAN}
Zaiqiao Meng, Shangsong Liang, Hongyan Bao, and Xiangliang Zhang.
\newblock Co-embedding attributed networks.
\newblock In {\em Proceedings of the Twelfth ACM International Conference on
  Web Search and Data Mining}, pages 393--401. ACM, 2019.

\bibitem{CCEM}
Yupeng Luo, Shangsong Liang, and Zaiqiao Meng.
\newblock Constrained co-embedding model for user profiling in question
  answering communities.
\newblock In {\em Proceedings of the 28th ACM International Conference on
  Information and Knowledge Management}, pages 439--448, 2019.

\bibitem{AVAE}
Xiang Zhang, Lina Yao, and Feng Yuan.
\newblock Adversarial variational embedding for robust semi-supervised
  learning.
\newblock {\em arXiv preprint arXiv:1905.02361}, 2019.

\bibitem{HGAT}
Weijian Chen, Yulong Gu, Zhaochun Ren, Xiangnan He, Hongtao Xie, Tong Guo,
  Dawei Yin, and Yongdong Zhang.
\newblock Semi-supervised user profiling with heterogeneous graph attention
  networks.
\newblock In {\em Proceedings of the 28th International Joint Conference on
  Artificial Intelligence}, pages 2116--2122. AAAI Press, 2019.

\bibitem{deeper}
Qimai Li, Zhichao Han, and Xiao-Ming Wu.
\newblock Deeper insights into graph convolutional networks for semi-supervised
  learning.
\newblock In {\em Thirty-Second AAAI Conference on Artificial Intelligence},
  2018.

\bibitem{chen2020measuring}
Deli Chen, Yankai Lin, Wei Li, Peng Li, Jie Zhou, and Xu~Sun.
\newblock Measuring and relieving the over-smoothing problem for graph neural
  networks from the topological view.
\newblock In {\em Proceedings of the AAAI Conference on Artificial
  Intelligence}, volume~34, pages 3438--3445, 2020.

\bibitem{dropedge}
Yu~Rong, Wenbing Huang, Tingyang Xu, and Junzhou Huang.
\newblock Dropedge: Towards deep graph convolutional networks on node
  classification.
\newblock {\em arXiv preprint arXiv:1907.10903}, 2019.

\bibitem{GAN}
Ian Goodfellow, Jean Pouget-Abadie, Mehdi Mirza, Bing Xu, David Warde-Farley,
  Sherjil Ozair, Aaron Courville, and Yoshua Bengio.
\newblock Generative adversarial nets.
\newblock In {\em Advances in neural information processing systems}, pages
  2672--2680, 2014.

\bibitem{LP}
Xiaojin Zhu and Zoubin Ghahramani.
\newblock Learning from labeled and unlabeled data with label propagation.
\newblock Report CMU CALD tech report CMU-CALD-02-107, Carnegie Mellon
  University, 2002.

\bibitem{BLA}
Carl Yang, Lin Zhong, Li-Jia Li, and Luo Jie.
\newblock Bi-directional joint inference for user links and attributes on large
  social graphs.
\newblock In {\em Proceedings of the 26th International Conference on World
  Wide Web Companion}, pages 564--573. International World Wide Web Conferences
  Steering Committee, 2017.

\bibitem{repara}
Diederik~P Kingma and Max Welling.
\newblock Auto-encoding variational bayes.
\newblock {\em arXiv preprint arXiv:1312.6114}, 2013.

\bibitem{CFVAE}
Dawen Liang, Rahul~G Krishnan, Matthew~D Hoffman, and Tony Jebara.
\newblock Variational autoencoders for collaborative filtering.
\newblock In {\em Proceedings of the 2018 World Wide Web Conference}, pages
  689--698, 2018.

\bibitem{NVAE}
Arash Vahdat and Jan Kautz.
\newblock Nvae: A deep hierarchical variational autoencoder.
\newblock {\em Advances in Neural Information Processing Systems}, 33, 2020.

\bibitem{progressiveGAN}
Tero Karras, Timo Aila, Samuli Laine, and Jaakko Lehtinen.
\newblock Progressive growing of gans for improved quality, stability, and
  variation.
\newblock In {\em International Conference on Learning Representations}, 2018.

\bibitem{largeGAN}
Andrew Brock, Jeff Donahue, and Karen Simonyan.
\newblock Large scale gan training for high fidelity natural image synthesis.
\newblock {\em arXiv preprint arXiv:1809.11096}, 2018.

\bibitem{ALA}
Stanislav Pidhorskyi, Donald~A Adjeroh, and Gianfranco Doretto.
\newblock Adversarial latent autoencoders.
\newblock In {\em Proceedings of the IEEE/CVF Conference on Computer Vision and
  Pattern Recognition}, pages 14104--14113, 2020.

\bibitem{AVB}
Lars Mescheder, S~Nowozin, and Andreas Geiger.
\newblock Adversarial variational bayes: Unifying variational autoencoders and
  generative adversarial networks.
\newblock In {\em 34th International Conference on Machine Learning (ICML)},
  pages 2391--2400. PMLR, 2017.

\bibitem{cvae-gan}
Jianmin Bao, Dong Chen, Fang Wen, Houqiang Li, and Gang Hua.
\newblock Cvae-gan: fine-grained image generation through asymmetric training.
\newblock In {\em Proceedings of the IEEE international conference on computer
  vision}, pages 2745--2754, 2017.

\bibitem{ARVGA}
Shirui Pan, Ruiqi Hu, Guodong Long, Jing Jiang, Lina Yao, and Chengqi Zhang.
\newblock Adversarially regularized graph autoencoder for graph embedding.
\newblock {\em arXiv preprint arXiv:1802.04407}, 2018.

\bibitem{VAEGAN}
Xianwen Yu, Xiaoning Zhang, Yang Cao, and Min Xia.
\newblock Vaegan: a collaborative filtering framework based on adversarial
  variational autoencoders.
\newblock In {\em Proceedings of the 28th International Joint Conference on
  Artificial Intelligence}, pages 4206--4212. AAAI Press, 2019.

\bibitem{AGC}
Zhiqiang Tao, Hongfu Liu, Jun Li, Zhaowen Wang, and Yun Fu.
\newblock Adversarial graph embedding for ensemble clustering.
\newblock In {\em Proceedings of the 28th International Joint Conference on
  Artificial Intelligence}, pages 3562--3568. AAAI Press, 2019.

\bibitem{vae-gcn}
Thomas~N Kipf and Max Welling.
\newblock Variational graph auto-encoders.
\newblock {\em arXiv preprint arXiv:1611.07308}, 2016.

\bibitem{InfoNCE}
Aaron van~den Oord, Yazhe Li, and Oriol Vinyals.
\newblock Representation learning with contrastive predictive coding.
\newblock {\em arXiv preprint arXiv:1807.03748}, 2018.

\bibitem{innerproduct}
Michael Tschannen, Josip Djolonga, Paul~K Rubenstein, Sylvain Gelly, and Mario
  Lucic.
\newblock On mutual information maximization for representation learning.
\newblock {\em arXiv preprint arXiv:1907.13625}, 2019.

\bibitem{GAT}
Petar Veli{\v{c}}kovi{\'c}, Guillem Cucurull, Arantxa Casanova, Adriana Romero,
  Pietro Lio, and Yoshua Bengio.
\newblock Graph attention networks.
\newblock {\em arXiv preprint arXiv:1710.10903}, 2017.

\bibitem{FB100}
Ryan~A. Rossi and Nesreen~K. Ahmed.
\newblock The network data repository with interactive graph analytics and
  visualization.
\newblock In {\em AAAI}, 2015.

\bibitem{snap}
Jure Leskovec and Andrej Krevl.
\newblock {SNAP Datasets}: {Stanford} large network dataset collection.
\newblock \url{http://snap.stanford.edu/data}, June 2014.

\bibitem{semi-gcn}
Thomas~N Kipf and Max Welling.
\newblock Semi-supervised classification with graph convolutional networks.
\newblock {\em arXiv preprint arXiv:1609.02907}, 2016.

\bibitem{eval}
Chuhan Wu, Fangzhao Wu, Junxin Liu, Shaojian He, Yongfeng Huang, and Xing Xie.
\newblock Neural demographic prediction using search query.
\newblock In {\em Proceedings of the Twelfth ACM International Conference on
  Web Search and Data Mining}, pages 654--662, 2019.

\bibitem{evalequation}
Donghui Wang, Yanchun Liang, Dong Xu, Xiaoyue Feng, and Renchu Guan.
\newblock A content-based recommender system for computer science publications.
\newblock {\em Knowledge-Based Systems}, 157:1--9, 2018.

\end{thebibliography}

\end{document}